\documentclass[11pt,a4paper]{article}
\usepackage[hyperref]{conll2018}
\usepackage{times}
\usepackage{latexsym}

\usepackage{url}
\usepackage{algorithm}
\usepackage{algorithmic}

\usepackage{amsmath}
\usepackage{amsfonts}
\usepackage{multirow}
\usepackage{makecell}
\usepackage{graphicx}
\aclfinalcopy 

\title{Chinese Poetry Generation with a Salient-Clue Mechanism}

\author{
$\text{Xiaoyuan Yi}^{1,2,3}$, $\text{Ruoyu Li}^5$,
$\text{Maosong Sun}^{1,2,4}$\thanks{\;\;Corresponding author: sms@mail.tsinghua.edu.cn.}
\\ 
$^1$Department of Computer Science and Technology, Tsinghua University \\
$^2$Institute for Artificial Intelligence, Tsinghua University \\
$^3$State Key Lab on Intelligent Technology and Systems, Tsinghua University\\
$^4$Beijing Advanced Innovation Center for Imaging Technology, Capital Normal University \\
$^5$6ESTATES PTE LTD, Singapore\\
{\tt yi-xy16@mails.tsinghua.edu.cn,} {\tt liruoyu@6estates.com},\\ {\tt sms@mail.tsinghua.edu.cn}
}

\date{}

\begin{document}
\maketitle
\begin{abstract}
As a precious part of the human cultural heritage, Chinese poetry has influenced people for generations. Automatic poetry composition is a challenge for AI. In recent years, significant progress has been made in this area benefiting from the development of neural networks. However, the coherence in meaning, theme or even artistic conception for a generated poem as a whole still remains a big problem. In this paper, we propose a novel Salient-Clue mechanism for Chinese poetry generation. Different from previous work which tried to exploit all the context information, our model selects the most salient characters automatically from each so-far generated line to gradually form a salient clue, which is utilized to guide successive poem generation process so as to eliminate interruptions and improve coherence. Besides, our model can be flexibly extended to control the generated poem in different aspects, for example, poetry style, which further enhances the coherence. Experimental results show that our model is very effective, outperforming three strong baselines.
\end{abstract}

\section{Introduction}
As a fascinating literary form starting from the Pre-Qin Period, Chinese poetry has influenced people for generations and thus influenced Chinese culture and history in thousands of years. Poets often write poems to record interesting events and express their feelings. In fact, the ability to create high-quality poetry has become an indicator of knowledge, wisdom and elegance of a person in China.
\begin{figure}
\centering
\includegraphics[scale=0.36]{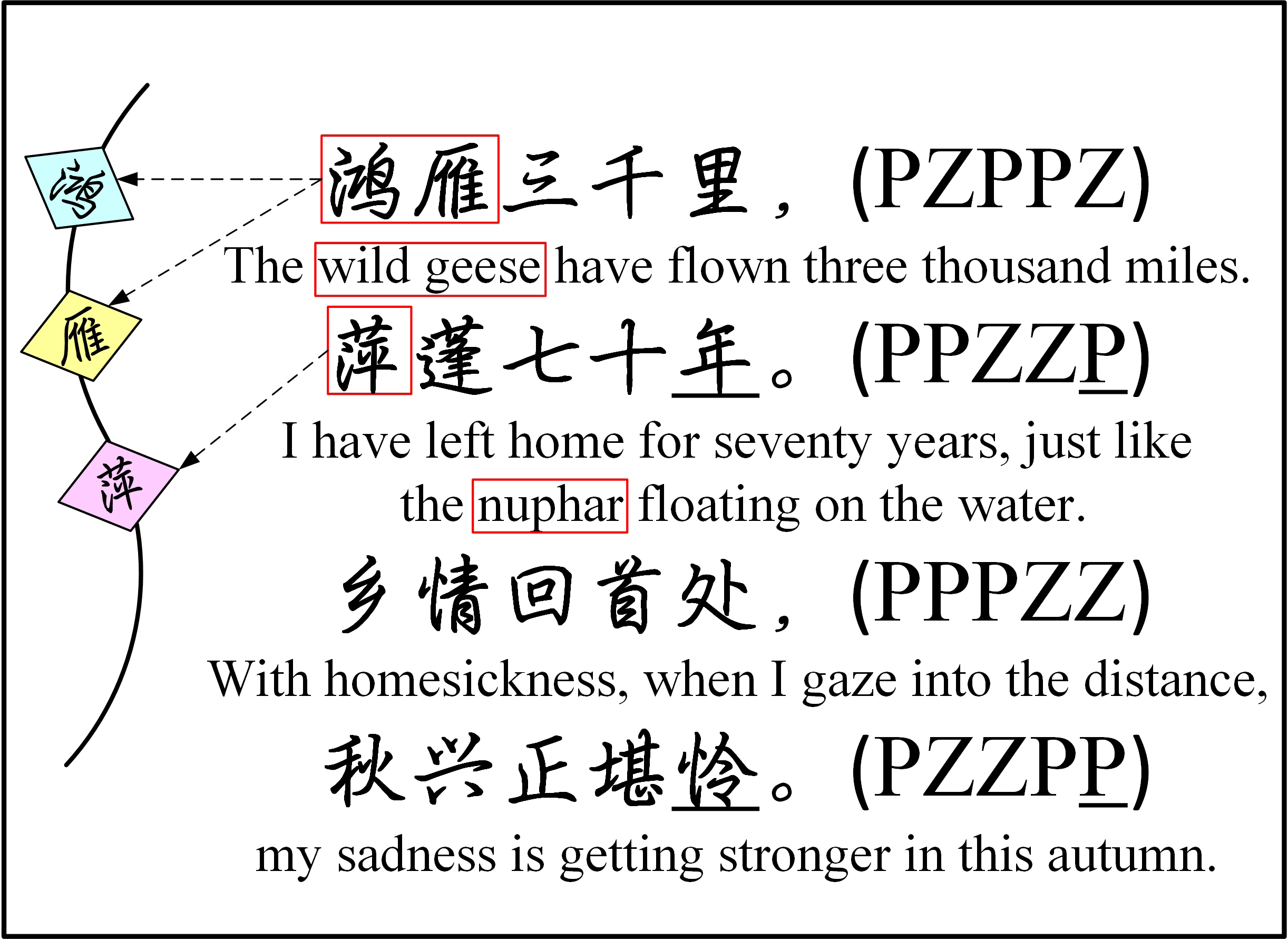}
\caption{A \emph{Wujue} generated by our model. The tone of each character is given in parentheses, where P and Z represent Ping (level tone) and Ze (oblique tone) respectively. Rhyming characters are underlined. The left part is an artistic illustration of the salient clue.}
\label{fig1}
\end{figure}

Generally, a Chinese poem should meet two kinds of requirements. One is from the perspective of \emph{form}: it must obey some structural and phonological rules strictly. For example (as shown in Figure \ref{fig1}), quatrain (\emph{Jueju} in Chinese), one of the most popular types of Chinese poetry, contains four lines with each consisting of five or seven characters (called \emph{Wujue} and \emph{Qijue} respectively); characters with particular tone must be in particular positions to make the poem cadenced and full of rhythmic beauty; and, the last character of the first (optional), second and fourth lines must rhyme. The other one is from the perspective of \emph{content}, concerning: (1) if each line of the poem is adequate syntactically and semantically; (2) if the association between two adjacent lines is reasonable; and (3) if the poem as a whole is coherent in meaning, theme or even in artistic conception. Obviously, the second requirement is much more complicated and difficult than the first one.

In this paper, we investigate on automatic Chinese poetry generation, with emphasis on quatrains. We believe the \emph{form} requirement is comparatively easy for a computer to deal with by some constraint checking.  For the \emph{content} requirement, point (1) and (2) can be also handled well owing to the use of powerful sequence-to-sequence neural networks \cite{Sutskever:14}, which are capable of producing well-formed target sentence given a source sentence. A challenging problem which remains unresolved for researchers is the point (3), where inter-lines associations are `global' throughout a poem, rather than `local' in point (2). The relevant experience tells us this is a major reason for the distinct gap between computer-generated poems and those written by poets. In fact, most previous models don't tackle this problem well and will produce incoherences and inconsistencies in generated poems.

Inter-lines coherence is the main concern of this paper. Intuitively, there should be a clear clue to keep the theme of a poem consistent. However, setting a fixed pattern of the clue in advance, e.g, pre-determining keywords for each line, may lose the flexibility and imagination, which are essential for poetry. When writing a poem, human poets will focus on some salient parts of the context to ignore distractions and create relevant content. During this process, poets gradually build a salient clue (or framework) of the poem \cite{Zhangy:15}, allowing not only coherence but also some flexibility.

Inspired by this, we propose a novel Salient-Clue Mechanism for poetry generation. Different from previous models which tried to exploit all the context, our model chooses a few salient characters out of each previously generated line, forming a vital clue for generating succeeding lines, so as to maintain the coherence of the whole poem to the maximum extent. In addition, owing to the flexible structure of our model, extra useful information (e.g., the user intent and poetry style) can be incorporated with the salient clue to control the generation process, further enhancing coherence.

Contributions of this work are as follows:
\begin{itemize}
\item To the best of our knowledge, we first propose to utilize the salient partial context to guide the poetry generation process. 
\item We extend our model to combine user intent and control the style of generated poems, which further enhance coherence.
\item Experimental results show that our model outperforms three strong baselines.
\end{itemize}

\section{Related Work}
The research on automatic poetry generation has lasted for decades. The early approaches are based on rules and templates, such as the ASPERA system \cite{Gervas:01} and Haiku system \cite{Wu:09}. Genetic algorithms are exploited to improve the quality of generated poems \cite{Manurung:03,Levy:01}. Other approaches are also tried, for instance, Yan et al. \shortcite{Yan:13} adopt the automatic summarization method. Following the work that successfully applies the Statistical Machine Translation approach (SMT) to the task of Chinese classical couplets generation \cite{Jiang:08}, He et al. \shortcite{He:12} further extend SMT to Chinese classical poetry generation. 

In recent years, a big change in research paradigm occurred in this field, that is, the adoption of neural network-based approaches, which have shown great advantages in both English poetry \cite{Hopkins:17,Ghazvininejad:17} and Chinese poetry generation, as well as other generation tasks. Context coherence is essential for text generation. In some related tasks, researchers have taken a step towards this goal, for example, the discourse Neural Machine Translation (NMT) \cite{Tiedemann:17,Maruf:17,Jean:17}. For poetry generation, some neural models have also recognized the importance of poem coherence. The fundamental issue here is how to define and use the context of a poem properly. 

Zhang and Lapata \shortcite{Zhang:14} first propose to generate Chinese poems incrementally with Recurrent Neural Network (RNN), which packs the full context into a single vector by a Convolutional Sentence Model. To enhance coherence, their model needs to be interpolated with two SMT features, as the authors state. Yan \shortcite{Yan:16} generates Chinese quatrains using two RNNs with an iterative polishing schema, which tries to refine the poem generated in one pass for several times. Yi et al. \shortcite{Yi:17} utilize neural Encoder-Decoder with attention mechanism \cite{Bahdanau:15} and trains different models to generate lines in different positions of a poem. Wang et al. \shortcite{Wang:16} propose a two-stage Chinese classical poetry generation method which at first plans the sub-keywords of the poem, then generates each line sequentially with the allocated sub-keyword. 
However, the beforehand extracted planning patterns bring some explicit constraints, which may take a risk of losing some degree of flexibility as discussed in Section 1.

These neural network-based approaches are very promising, but there is still large room for exploration. For instance, whether packing the full context into a single vector really represents the `full' context as well as expected? Can we do better to represent the inter-lines context more properly in pursuing better coherence of the entire generated poem? Our work tries to respond to these questions.

\section{Model Design}
\label{sec3}
We begin by formalizing our problem. Suppose a poem P consists of n lines, $P=L_{1}L_{2} \ldots L_{n}$. Given previous i-1 lines $L_{1:i-1}$, we need to generate the i-th line which is coherent with the context in theme and meaning. Since our model and most baselines are based on a powerful framework first proposed in NMT, that is, the Bidirectional LSTM \citep{Schuster:97,Hochreiter:97} Encoder-Decoder with attention mechanism \cite{Bahdanau:15}, we first denote X a line in Encoder, $X=(x_1x_2 \ldots x_T)$, and Y a generated line in Decoder, $Y=(y_1y_2 \ldots y_T)$.  $T$ is the length of a line. $h_t$ and ${h_t}^{'}$ represent the Encoder and Decoder hidden states respectively. $emb(y_{t-1})$ is the word embedding of $y_{t-1}$. The probability distribution of each character to be generated in the i-th line is calculated by:\footnote{For brevity, we omit biases and use $h_t$ to represent the combined state of bidirectional LSTM Encoder.}
\begin{align}
& h^{'}_t = LSTM(h^{'}_{t-1}, emb(y_{t-1}), c_t), \\
& p(y_t|y_{1:t-1},L_{1:i-1}) = g(h^{'}_t, emb(y_{t-1}), c_t, v),
\end{align}

where $g$ is a normalization function, softmax with a maxout layer \cite{Goodfellow:13} in this paper.  $y_{1:t-1}$ means $y_1, \ldots ,y_{t-1}$ (similar to $L_{1:i-1}$). $c_t$ is the local context vector in attention mechanism. $v$ is a global context vector. To avoid confusion, in the remainder of this paper when it comes to the word `context', we all mean the global context, that is, so-far generated lines $L_{1:i-1}$.

Now the key point lies in how to represent and utilize the context for the sake of better coherence. Before presenting the proposed method, we first introduce two basic formalisms of utilizing full context.

\subsection{Basic Models}
\label{subsec31}
\subsubsection{nLto1L}
We call the first formalism \textbf{nLto1L}, where poetry generation is regarded as a process similar to machine translation. The difference is that the pair in the parallel corpus for NMT models, is changed to the pair of $<$preceding lines in a poem, a line in a poem$>$ here, which is semantically related rather than semantically equivalent.
\begin{figure}
\centering
\includegraphics[scale=0.23]{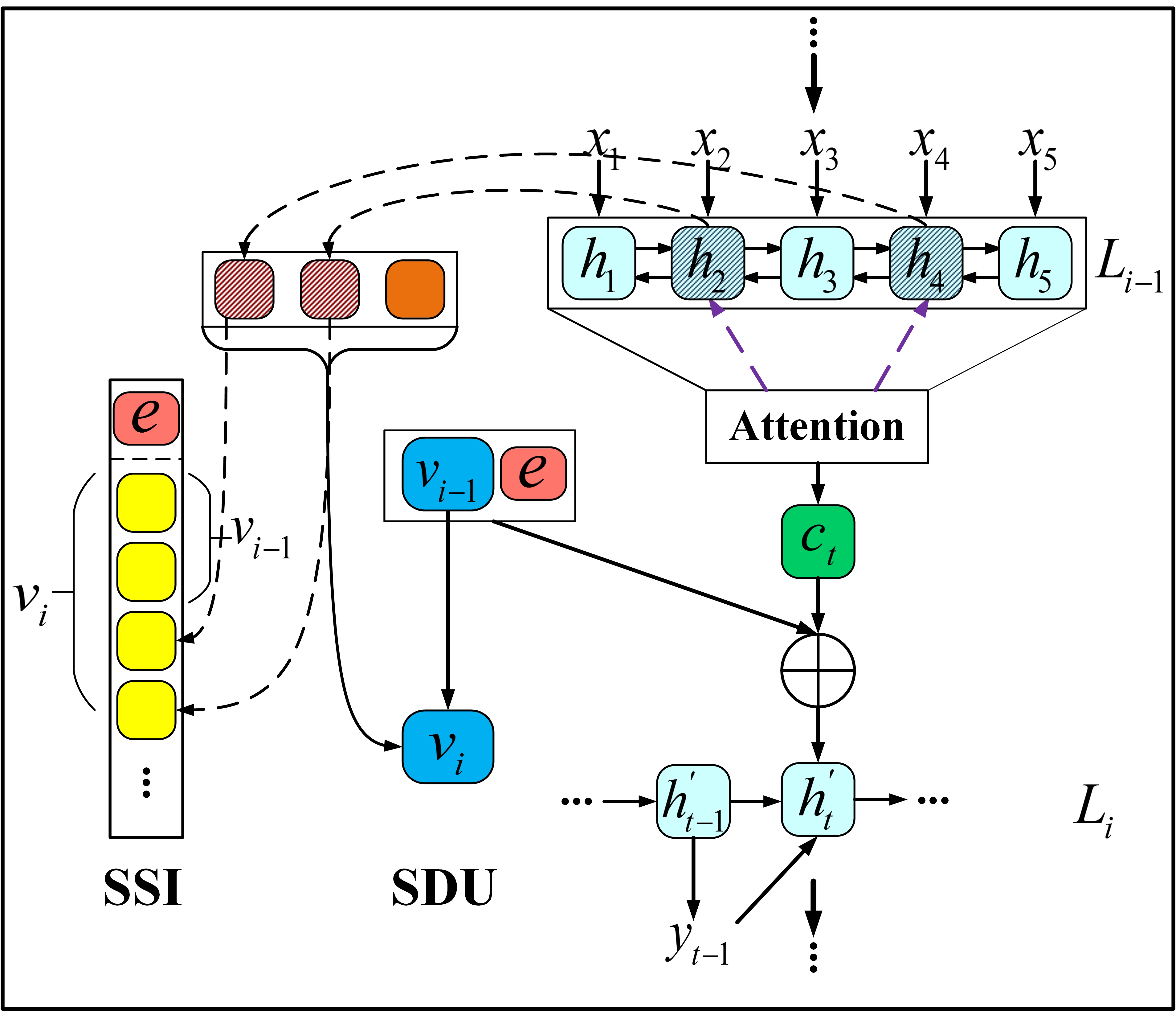}
\caption{A graphical illustration of the proposed Salient-Clue mechanism. $v_i$ is the salient-clue vector and $e$ is the extension vector. We design two strategies for updating the salient clue. SDU: $v_i$ is kept at the same size; SSI: the size of $v_i$ increases during the generation process.}
\label{fig2}
\end{figure}

The `n' in nLto1L means at most n preceding lines are concatenated as a long sequence and used simultaneously in Encoder, corresponding to the preceding-lines-in-poem part in the pair, to generate a line in Decoder.  In this case, the context is captured by $c_t$ without extra $v$. \cite{Wang:16} and \cite{Yi:17} both belong to this formalism.

The nLto1L formalism is effective, but it has two drawbacks. For one thing, as `n' increases in nLto1L, more global context can be exploited explicitly by attention, but the number of training pairs decreases, which hurts the generalization performance. For instance, from each quatrain, only one 3Lto1L pair (but two 2Lto1L pairs) can be extracted. For another, when the input sequence is too long, the performance of NMT models will still degrade, even with an attention mechanism \cite{P16-1159}. We find this problem more prominent in poetry generation, since attention may fail to capture all important parts of the context. Our preliminary experiment shows that, regarding generating the fourth line, both for Yi's model and Wang's model, more than 70\% of the top three attention values fall into just the third line area and thus neglect other lines, which validates our assumption.
\begin{figure}
\centering
\includegraphics[scale=0.25]{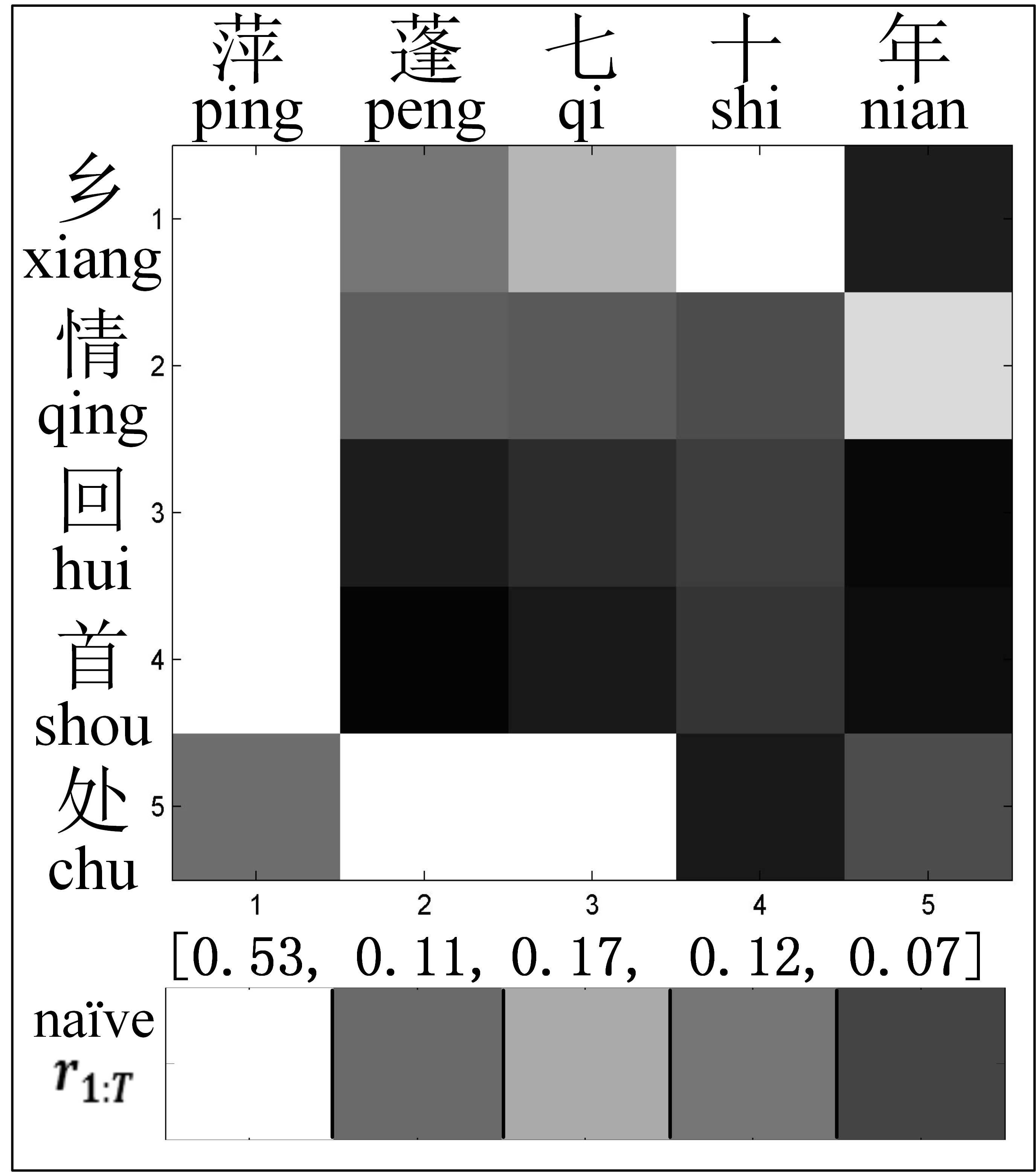}
\caption{An example of calculating the saliency score of each character (in the x-axis) from the attention matrix (0:black, 1:white), in the naive Salient-Clue. The scores are normalized to interval [0,1] here.}
\label{fig3}
\end{figure}
\subsubsection{Packing Full Context}
Another formalism is to pack the full context into a single vector $v$, which is used to generate successive lines \cite{Zhang:14, Yan:16}. Usually, $v$ is updated by the vector of each generated line in a poem. This formalism is not as powerful as we expected. There is still much room for improvement. A single vector doesn't have enough capacity to store all context. Moreover, meaningful words and noises (e.g., stop words) are mixed in one vector, which results in the implicit and indiscriminate utilization of the context.

\subsection{The Proposed Salient-Clue Mechanism}
\label{subsec32}
As discussed, using the full context directly cannot necessarily lead to the best performance. It becomes clear that we still need to develop a new mechanism to exploit the context in a proper way. Our design philosophy is ignoring the uninformative parts (e.g., stop words) and using some salient characters in context to represent the full context and form a salient clue, which is used to guide the generation process. Following this idea, we propose our Salient-Clue Model.
\begin{algorithm}[t]  
\caption{Saliency Selection Algorithm}   
\label{alg1}   
\begin{algorithmic}[1] 
\REQUIRE The saliency scores of characters in the preceding line, $r_{1:T}$; $K$;\\  
\ENSURE The number of finally selected salient characters, $N$; The indices of selected characters in the preceding line, $m_{1:N}$;  

\STATE Calculate the mean value of $r_{1:T}$, $avg$;   
\STATE Calculate the standard deviation of $r_{1:T}$, $std$; 
\STATE Get sorted indices $i_{1:T}$ in descending order of $r_{1:T}$;    
\STATE $k = 1$; $val =  avg + 0.5* std$;
\WHILE {$(r_{i_k} \geq val ) \quad and \quad ( k \leq K)$}
\STATE $m_k = i_{k}$; $val = val * 0.618$ (the golden ratio); $k = k+1$; 
\ENDWHILE 
\STATE $N = k-1$;
\RETURN $N$, $m_{1:N}$; 
\end{algorithmic}
\end{algorithm}
\subsubsection{Naive Salient-Clue}
\begin{figure}[t]
\centering
\includegraphics[scale=0.25]{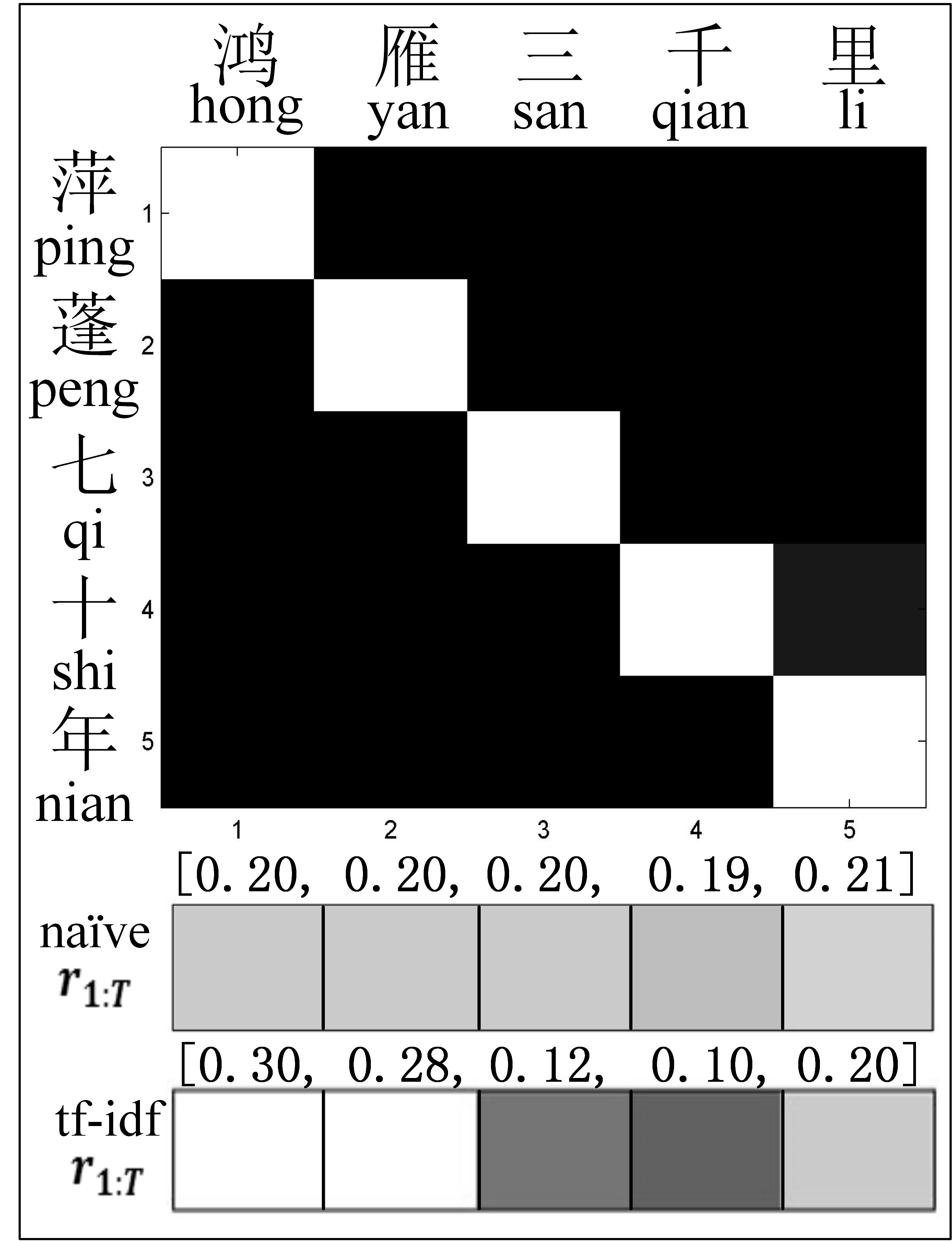}
\caption{The comparison of saliency scores obtained by the naive model and the tf-idf weighting improved model: an example.}
\label{fig4}
\end{figure}

As illustrated in Figure \ref{fig2}, to generate $L_i$, our model uses standard attention mechanism to exploit $L_{i-1}$ so as to capture short-distance relevance. And it utilizes a salient-clue vector, $v_i$, to exploit long-distance context. After generating $L_i$, our model selects up to K (K is 2 for \emph{Wujue} and 3 for \emph{Qijue} in this work) most salient characters from $L_{i-1}$ according to the attention values, and uses their corresponding Encoder hidden states to update the salient clue vector $v_i$. Thanks to the bidirectional LSTM, even if we only focus on part of the context characters, the information of those unselected won't be lost completely.

Concretely, let $A$ denote the attention alignment matrix in the attention mechanism \cite{Bahdanau:15} between the preceding line $L_{i-1}$ and the current generated line $L_i$. We calculate the \textbf{saliency score} of j-th character in $L_i$, $r_j$, by:
\begin{align}
& r_j = \frac{\sum_{i=1}^{T}A_{ij}}{\sum_{j^{'}=1}^{T}\sum_{i=1}^{T}A_{ij^{'}}},
\end{align}

where $A_{ij}$ is the element in i-th row and j-th column of $A$. Figure \ref{fig3} depicts an example. The most salient character is \emph{``ping''} (nuphar, a kind of plant, a symbol of loneliness) and the second one is \emph{``qi''} (seven), according to their saliency scores r(\emph{ping}) = 0.53 and r(\emph{qi}) = 0.17. 

The character \emph{``ping''} is very informative for the generated poem but \emph{``qi''} isn't, as signaled by the sharp distinction between their saliency scores. So we design the \textbf{Saliency Selection Algorithm \ref{alg1}} to further filter out characters with quite low saliency scores, like \emph{``qi''} here. We define this algorithm as a function $SSal(r_{1:T},K)$, which takes the saliency scores and the maximum number of selected characters as inputs and outputs the number of finally selected characters and their indices. Then we update the salient-clue vector $v_i$ as follows:
\begin{align}
& N, m_{1:N} = SSal(r_{1:T},K), \\
& s = \frac{ \sum_{k=1}^{N}r_{m_k} * h_{m_k} } {\sum_{k^{'}}^N r_{m_{k^{'}}}}, \\
& v_{i} = \sigma(v_{i-1}, s), v_0 = \vec{0},
\end{align}

where $\sigma$ is a non-linear layer. $v_{i-1}$ is used to predict next character to be generated by formula (2). Please note that in formula (4), N may be smaller than K since we want to further ignore relatively less salient characters even though they are already in the list of the K most salient ones.

By the generation process presented above, each generated line is guided by the salient clue and therefore is coherent with it. Meanwhile, informative parts of each generated line are selected and maintained in the salient clue. As a result, the salient clue vector always keeps a coherent information flow, playing a role of a dynamically and incrementally built framework (skeleton) for the generated poem.

\subsubsection{TF-IDF Weighted Attention}
Observe an example in Figure \ref{fig4}. The scores $r_{1:T}$ given by the Naive Salient-Clue are very close to each other, not distinguishable in saliency. To cope with this, we further take into account the importance of characters both in the preceding line and the current generated line, by the traditional tf-idf scheme in Information Retrieval: 
\begin{align}
& r_j = [(\mathbf{w}_{out} * A) \odot \mathbf{w}_{in}]_j,
\end{align}

where $\odot$ is element-wise multiplication and $[\cdot]_j$ is the j-th element in a vector. $\mathbf{w}_{in} \in \mathbb{R}^{1*T}$ is the tf-idf vector of preceding (input) line and the i-th element of it is the tf-idf value of i-th character. Similarly, $\mathbf{w}_{out}$ is the tf-idf vector of the current generated (output) line. Elements in $\mathbf{w}_{in}$ and $\mathbf{w}_{out}$ are normalized to [0,1].

As shown in Figure \ref{fig4}, by tf-idf weighting, two informative characters \emph{``hong yan''} (wild goose, a symbol of autumn) are selected correctly, which leads to the generation of word \emph{``qiu xing''} (sadness in autumn) in the fourth line in Figure \ref{fig1}.

\subsubsection{Two Strategies for Salient-Clue}
As shown in Figure \ref{fig2}, we use two strategies to form and utilize the salient-clue vector $v_i$. The first is called \textbf{Saliency Dynamic Update (SDU)} by formula (5) and (6), which means that hidden states of selected salient characters are packed into $v_i$. Thus $v_i$ is kept at the same size and is updated dynamically after each line is generated. 

The second one is the concatenation of these hidden states:
\begin{align}
& v_i = [v_{i-1};h_{m_1};...;h_{m_N}],
\end{align}

where $[;]$ means vector concatenation. The size of $v_i$ will increase in the generation process. We call this \textbf{Saliency Sensitive Identity (SSI)}, because the identity of each hidden state is kept independent, without being merged as one.

\subsection{Extensions of Salient-Clue}
\label{subsec33}
Above we design different methods to select the salient partial context. Since the proposed model is quite flexible, aside from the selected characters, other information can be also utilized to form the salient clue, so as to further improve coherence. In this paper, we tried and evaluated two extensions: user intent and poetry style. This extra information is vectorized as an extension vector $e$ and then concatenated to the salient clue vector:
\begin{align}
\begin{split}
p(y_t|y_{1:t-1},L_{1:i-1}) & =  \\ 
& g(h^{'}_t, emb(y_{t-1}), c_t, [v_{i-1};e]).
\end{split}
\end{align}

\textbf{Intent Salient-Clue.} The poem is generated with a user intent keyword. Taking user intent into account can prevent later generated lines diverging from earlier generated ones in a poem. In detail, we feed the keyword into Encoder, then vector $e$ is calculated by a non-linear transformation of the average of their hidden states.

\textbf{Style Salient-Clue.} The style of generated poems can also benefit coherence. Here we simply use a style embedding as the vector $e$, which provides a high-level indicator of style and is learned during the training process. It is noteworthy that Zhang et al. \shortcite{Zhang:17} achieve style transfer with the help of an external memory, which stores hundreds of poems (thus thousands of hidden states). By contrast, our Style extension is simpler but still achieves comparable performance.
\begin{table}
\small
\centering
\begin{tabular}{|c||l|l|l|}

\Xhline{1.2pt}
& \bf Models & \bf \emph{Wujue} & \bf \emph{Qijue} \\
\Xhline{1.2pt}

\multirow{2}{*}{\bf Different} & Planning  & 0.460 &  0.554 \\ 
\cline{2-4}

& iPoet & 0.502 &  0.591  \\
\cline{2-4}

\multirow{2}{*}{\bf Models} & seq2seqPG & 0.466 & 0.620    \\
\cline{2-4}

& SC & \bf 0.532 &  \bf 0.669 \\
\Xhline{1.2pt}

\multirow{2}{*}{\bf Different} & naive-TopK-SDU & 0.442 &  0.608  \\
\cline{2-4}

& naive-SSal-SDU & 0.471 &  0.610  \\
\cline{2-4}

\multirow{1}{*}{\bf Strategies}& tfidf-SSal-SDU & \bf 0.533 &  0.648  \\
\cline{2-4}

\multirow{2}{*}{\bf of SC } & tfidf-SSal-SSI & 0.530 & 0.667  \\
\cline{2-4}

& tfidf-SSal-SSI-intent & 0.532 & \bf 0.669  \\
\cline{2-4}

\Xhline{1.2pt}

\end{tabular}
\caption{BLEU evaluation results. The scores are calculated by the multi-bleu.perl script.}
\label{tab1}
\end{table}
\section{Experiments and Evaluations}
\label{sec4}
\subsection{Data and Setups}
Our corpus contains 165,800 poems (half \emph{Wujue} and half \emph{Qijue}). We use 4,000 of them for validation, 4,000 for testing and other ones for training. From each poem, a keyword is extracted by tf-idf.

The sizes of word embedding, hidden state, saline-clue vector, intent vector and style embedding are set to 256, 512, 512, 128 and 64 respectively. For SSI, to reduce the model size, we map each hidden state to a 100-d vector by a non-linear transformation. Encoder and Decoder share the same word embedding. All different strategies of Salient-Clue are used both in training and generation. The optimization objective is standard cross entropy errors of the predicted character distribution and the actual one. Adam \cite{Kingma:14} with shuffled mini-batches (batch size 64) is used. Then we use beam search (beam size 20) to generate each poem, with a keyword as input. For fairness, all baselines use the same configuration.

For Style Salient-Clue model, we first use LDA \cite{Blei:03} to train the whole corpus (with 15 topics). We select three main styles in Chinese poetry: Pastoral, Battlefield and Romantic and find the corresponding topics manually. Then all poems are labeled by LDA inference. We select 5,000 poems for each style, together with other 5,000 non-style poems (20,000 in total), to fine-tune a pre-trained normal Salient-Clue model.

\subsection{Models for Comparisons}
We compare: \textbf{iPoet} \cite{Yan:16}, \textbf{seq2seqPG} \cite{Yi:17}, \textbf{Planning} \cite{Wang:16}, \textbf{SC} (our model, tfidf-SSal-SSI-intent, which is the best configure under BLEU evaluation), \textbf{Style-SC} (the style extension of our model) and \textbf{Human} (poems created by human poets). We choose these three previous models as our baselines, because they all achieve satisfactory performance and the authors have done thorough comparisons with other models, such as RNNPG \cite{Zhang:14} and SMT \cite{He:12}, and prove that their models outperform baselines. Besides, the three models can be classified into the two formalisms in Section 3.1.
\begin{table*}[htp]
\small
\centering
\begin{tabular}{|l|l|l|l|l|l|l|l|l|l|l|}

\Xhline{1.2pt}
\multirow{2}{*}{\bf Models} & \multicolumn{2}{c|}{ \bf Fluency} & \multicolumn{2}{c|}{\bf Coherence} & \multicolumn{2}{c|}{\bf Meaningfulness}  & \multicolumn{2}{c|}{\bf Poeticness} & \multicolumn{2}{c|}{\bf Entirety} \\

\cline{2-11}
& \emph{Wujue} & \emph{Qijue} & \emph{Wujue} & \emph{Qijue} & \emph{Wujue} & \emph{Qijue} & \emph{Wujue} & \emph{Qijue} & \emph{Wujue} & \emph{Qijue} \\

\Xhline{1.2pt}
Planning & 2.56 & 2.84 & 2.50 & 2.64 & 2.49 & 2.64 & 2.59 & 2.88 & 2.39 & 2.66\\
\hline
iPoet    & 3.13 & 3.45 & 2.89 & 2.91 & 2.60 & 2.80 & 2.79 & 3.05 & 2.54 & 2.85\\
\hline
seq2seqPG& 3.54 & 3.65 & 3.31 & 3.16 & 3.15 & 3.01 & 3.26 & 3.29 & 3.06 & 3.08\\
\hline
SC       & 4.01$^{**}$ & 4.04$^{**}$ & 3.85$^{**}$ & 3.86$^{**}$ & 3.55$^{**}$ & 3.63$^{**}$ &\bf 3.74$^{**}$ &\bf 3.69$^{*}$ & 3.63$^{**}$ &\bf 3.70$^{**}$\\
\hline
Style-SC &\bf 4.03$^{**}$ &\bf 4.16$^{**}$ & \bf 3.90$^{**}$ & \bf 4.01$^{**}$ &\bf 3.68$^{**}$ &\bf 3.75$^{**}$ & 3.61$^{*}$ & 3.68$^{*}$ &\bf 3.65$^{**}$ &\bf 3.70$^{**}$\\
\hline
Human    & 4.09 & 4.43 & 3.90 & 4.33$^{+}$ & 3.94 & 4.35$^{++}$ & 3.83 & 4.24$^{++}$ & 3.81 & 4.24$^{++}$ \\
\Xhline{1.2pt}
\end{tabular}
\caption{Human evaluation results. Diacritics * ($p < 0.05$) and ** ($p < 0.01$) indicates SC models significantly outperform the three baselines; + ($p < 0.05$) and ++ ($p < 0.01$) indicates Human is significantly better than all the five models. The Intraclass Correlation Coefficient of the four groups of scores is 0.596, which indicates an acceptable inter-annotator agreement.}
\label{tab2}
\end{table*}
\subsection{Evaluation Design}
To demonstrate the effectiveness of our model, we conduct four evaluations:
 
\textbf{BLEU Evaluation.} Following \cite{He:12,Zhang:14,Yan:16}, we use BLEU \cite{P02-1040} to evaluate our model. BLEU isn't a perfect metric for generated poems, but in the scenario of pursuing better coherence, it still makes sense to some extent. Because higher BLEU scores indicate that the model can generate more n-grams of ground-truth, which certainly have better coherence.

\textbf{Human Evaluation.} Following \cite{Manurung:03,Zhang:14}, we design five criteria: \textbf{Fluency} (are the lines fluent and well-formed?), \textbf{Coherence} (is the theme of the whole quatrain consistent?), \textbf{Meaningfulness} (does the poem convey some certain messages?), \textbf{Poeticness} (does the poem have some poetic features?), \textbf{Entirety} (the reader's general impression on the poem). Each criterion needs to be scored in a 5-point scale ranging from 1 to 5. 

We select 20 typical keywords and generate two quatrains (one \emph{Wujue} and one \emph{Qijue}) for each keyword using these models. For Human, we select quatrains containing the given keywords. Therefore, we obtain 240 quatrains (20*6*2) in total. We invite 12 experts on Chinese poetry to evaluate these quatrains, who are Chinese literature students or members of a poetry association. Experts are divided into four groups and required to focus on the quality as objectively as possible, even if they recognize the human-authored ones. Each group completes the evaluation of all poems and we use the average scores.

Style-SC is not suitable for BLEU, because we can't expect LDA to predict a correct style label by a short keyword. Thus Style-SC is only tested under Human Evaluation. We label each keyword with an appropriate style manually, which is used to guide the generation.

\textbf{Style Control Evaluation.} Poetry style is usually coupled with content. Not all keywords are compatible with every style. Therefore we select ten normal keywords without obvious style (e.g., moon and wind). We use SC to generate one poem and use Style-SC to generate three poems with the three specified styles. The experts are asked to identify the style of each poem from four options (Unknown, Battlefield, Romantic and Pastoral).

\textbf{Saliency Selection Evaluation.} The main idea of our method is to select the salient partial context to guide successive generation. To evaluate the reasonableness of selected characters, we randomly select 20 \emph{Wujue}s and 20 \emph{Qijue}s from the test set. Then three experts are asked to select up to K salient characters from each line. When experts have different opinions, they stop and discuss until reaching an agreement. \textbf{Jaccard similarity} is used to measure the overlap of human-selected characters and the model-selected ones.

\subsection{Evaluation Results and Discussion}
As shown in Table \ref{tab1}, our SC outperforms other models under BLEU evaluation. We also compare different strategies of SC. As we expected, tfidf-SC models outperform the naive ones, because tf-idf values lower the weights of uninformative characters. We also compare our SSal \ref{alg1} with TopK (just select top K salient characters) and SSal gets better results. Please note that, from naive-TopK-SDU to tfidf-SSal-SDU, BLEU scores are getting higher without any increase of model size (Table \ref{tab4}). SSI is better on \emph{Qijue}, but performs slightly worse than SDU on \emph{Wujue}. We use SSI for Human evaluation here, but SDU is more suitable for longer poetry, e.g., Chinese Song Iambics. Besides, the intent extension makes a little bit of improvement, not as prominent as we expected. We think the reason may lie in that the keyword selected by tf-idf can't accurately represent the user intent. Generally, the results show both for packing and concatenation formalisms, the proper utilization of partial salient context (SDU and SSI) can be better than the improper utilization of full context (Packing Full Context and nLto1L).

Table \ref{tab2} gives human evaluation results. SC and Style-SC achieve better results than other models and get close to Human, though there is still a gap. Especially on Coherence, our Style-SC gets the same score as Human for \emph{Wujue}. Moreover, Style-SC makes a considerable improvement on Coherence compared to SC (+0.05 for \emph{Wujue} and +0.15 for \emph{Qijue}), which demonstrates that consistent style can actually enhance the coherence, though it's not easy to predict an appropriate style automatically. Interestingly, Style-SC outperforms SC on most criteria, except for Poeticness. We believe this is mainly because that style control forces the model to always generate some style-related words, which limits the imagination and thus hurts the Poeticness.

Besides, as we can see, seq2seqPG outperforms other two baselines, but at the expense that it is three times the model size of iPoet (Table \ref{tab4}). Surprisingly, Planning gets the worst results. This is because that Planning uses a neural language model to generate the planned sub-keywords, which performs poorly on our corpus and thus hurts fluency and coherence.

\begin{figure}
\centering
\includegraphics[scale=0.25]{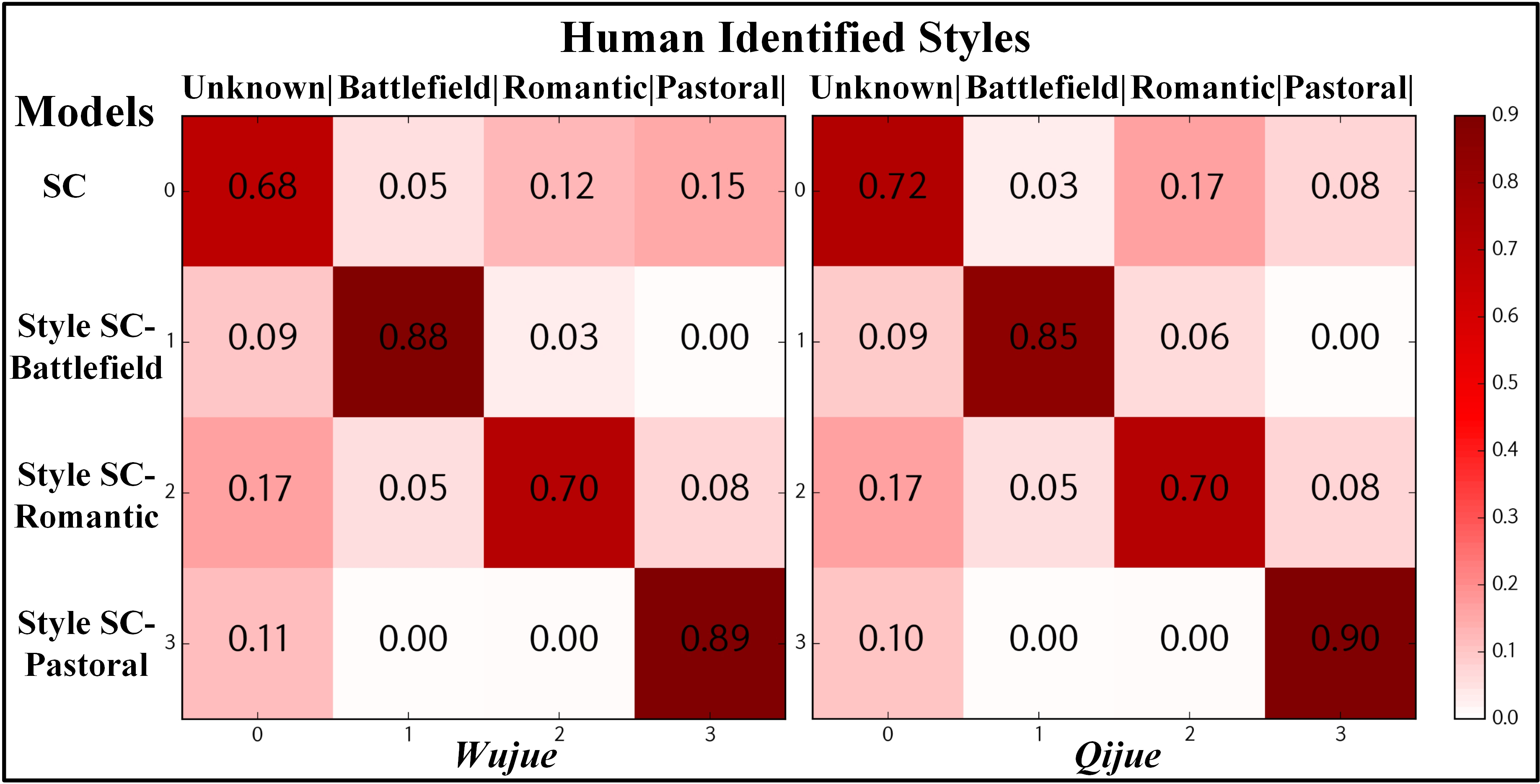}
\caption{Style control evaluation results. The values are ratios that generated poems are identified as different styles by human experts.}
\label{fig5}
\end{figure}

Figure \ref{fig5} gives style control evaluation results. Incorporating one style vector with the salient-clue, our Style-SC achieves comparable performance with \cite{Zhang:17}. The good results partially lie in that we only use those non-style keywords, such as moon, wind, water and so on, for this experiment. Empirically, transferring from a keyword with obvious style to arbitrary style is intractable, e.g., from the word `army' to Pastoral style, which may need more complicated model design. Even so, our results still show that rough style control is not as difficult as we thought. 

Table \ref{tab3} shows the effectiveness of our salient character selection methods. tfidf-SSal achieves about 50\% overlap of human-selected salient characters, which is notably higher than Random and tf-idf. With only attention values, naive-TopK performs worse than tf-idf on \emph{Wujue}. Combining tf-idf and SSal makes notable improvement.
\begin{table}
\centering
\begin{tabular}{|l|l|l|}

\Xhline{1.2pt}
\bf Models & \bf \emph{Wujue} & \bf \emph{Qijue}  \\
\Xhline{1.2pt}

Random & 0.271 &  0.247  \\
\hline

tf-idf & 0.417 &  0.378  \\
\hline

naive-TopK SC & 0.347 &  0.415  \\
\hline

naive-SSal SC &  0.431 & 0.441  \\
\hline

tfidf-SSal SC & \bf 0.525 & \bf 0.461  \\

\Xhline{1.2pt}
\end{tabular}
\caption{Saliency selection results. Random: randomly select K characters for three times and use the average Jaccard values. tf-idf: directly select K characters in terms of tf-idf, without SC. }
\label{tab3}
\end{table}
\subsection{Extra Comparisons and Case Study}
Besides the metrics above, we also compare innovation, model size and generation speed of different models. The innovation is measured by Jaccard similarity of generated poems (3,000 for each model). Intuitively, the basic requirement for innovation is that poems generated with different keywords should be different with each other. 

As shown in Table \ref{tab4}, SC makes a good balance on quality, innovation, generation speed and model size. The iPoet has the smallest size, but the generation is slow, since it may polish the poem for several times, costing more time than other one-pass-generation models. For SC, the use of tf-idf significantly improves innovation. Due to planned sub-keywords, Planning achieves the best innovation but the worst quality, which shows pursuing innovation takes the risk of abruptness and incoherence.

\begin{figure}
\centering
\includegraphics[scale=0.07]{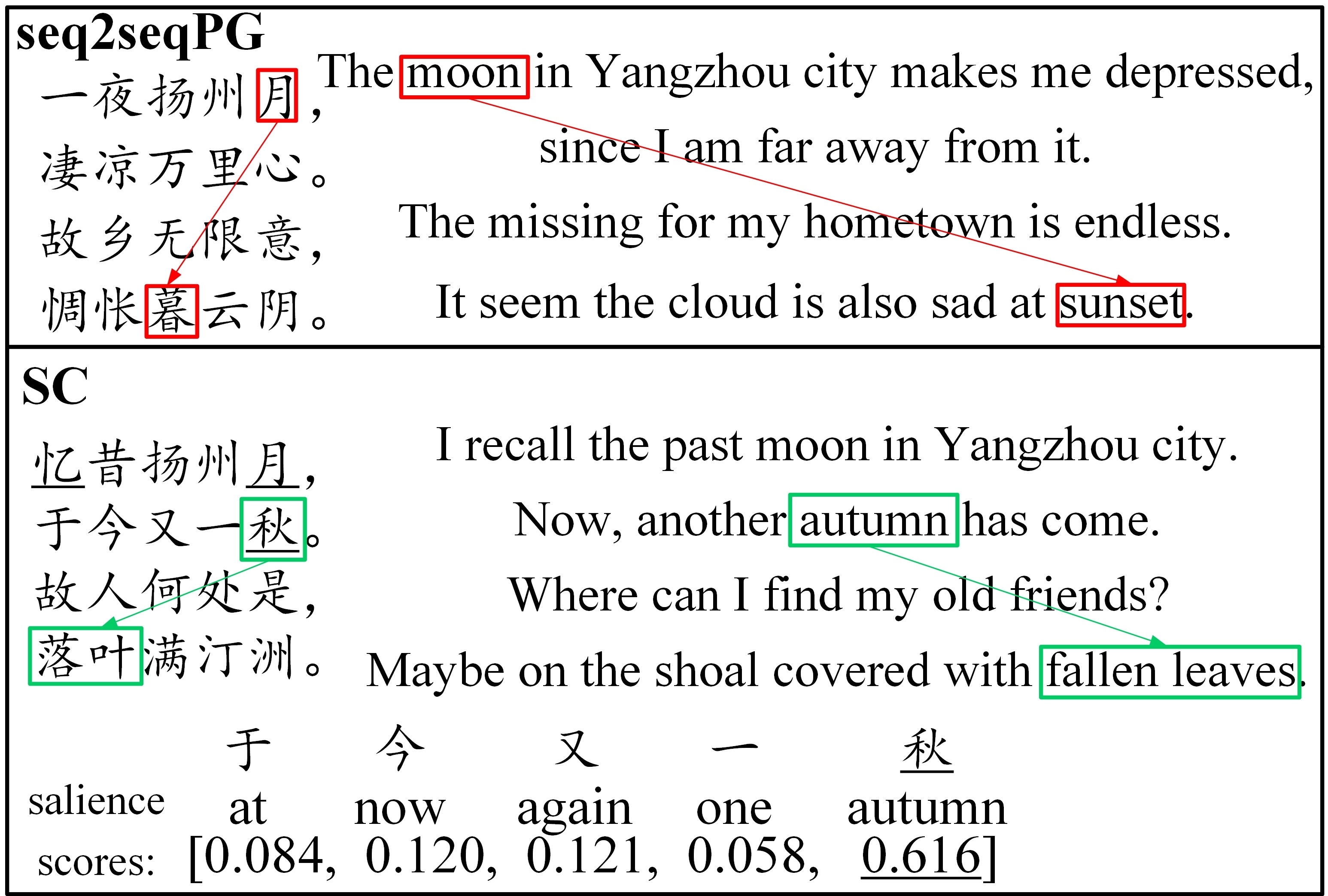}
\caption{Two \emph{Wujue}s generated with the same input. Green boxes and arrows show consistencies, and the red ones show inconsistencies. Automatically selected charcters by SC are underlined.}
\label{fig6}
\end{figure}

Figure \ref{fig6} shows two \emph{Wujue}s generated by seq2seqPG and SC respectively, with the same input ``Yangzhou City''. A word ``moon'' is generated by seq2seqPG in the first line, which determines the time (at night) of the whole poem. However, in the fourth line, seq2seqPG still generates an inconsistent word ``sunset''. For the poem generated by SC, the word ``autumn'' in the second line is selected for successive generation. As a result, a word ``fallen leaves'' is generated in the fourth line. Furthermore, in the second line, except for ``autumn'', other four uninformative characters, which have quite low saliency scores, are filtered out by \textbf{SSal} \ref{alg1} as shown at the bottom of Figure \ref{fig6}.
\begin{table}
\small
\centering
\begin{tabular}{|l|c|c|c|}

\Xhline{1.2pt}
\bf Models & \bf Innovation & \bf Param & \bf Speed \\
\Xhline{1.2pt}

Planning  & \bf 0.039 & 15.1 & 2.49\\
\cline{2-4}

iPoet & 0.044 & \bf 11.0  &  1.76\\
\cline{2-4}

seq2seqPG & 0.047 & 37.1 & 1.85 \\
\cline{2-4}

SC Model &  0.041 &   13.6 & \bf 1.57 \\
\Xhline{1.2pt}

naive-TopK-SDU & 0.058 &  13.7 & 1.49 \\
\cline{2-4}

naive-SSal-SDU & 0.056 &  13.7 & 1.51\\
\cline{2-4}

tfidf-SSal-SDU & 0.042 &  13.7 & 1.52 \\
\cline{2-4}

tfidf-SSal-SSI &  0.043 & \bf 13.1 & \bf 1.48  \\
\cline{2-4}

tfidf-SSal-SSI-intent & \bf 0.041 &  13.6 & 1.57 \\
\cline{2-4}

\Xhline{1.2pt}

\end{tabular}
\caption{Extra comparisons of different models. Innovation, Param (million parameters) and Speed (seconds per poem) are compared. The generation speed is tested on an Intel CORE i5 4-core CPU.}
\label{tab4}
\end{table}
\section{Conclusion and Future Work}
\label{sec5}
In this paper, we address the problem of the context coherence in poetry generation. How to properly treat the global context is a key factor to consider. We propose a Salient-Clue mechanism\footnote{Based on this work, we build an online poetry generation system, \textbf{Jiuge}: \url{https://jiuge.thunlp.cn}.}. Our model selects highly salient characters in preceding generated lines to form a representation of the so-far context, which can be considered as a dynamically and incrementally built framework, then uses it to guide the successive generation. Both automatic and human evaluations demonstrate that our model can effectively improve the global coherence of meaning, theme and artistic conception of generated poems. This implies the proper treatment of a partial context could be better than the improper treatment of the full context. 

Furthermore, our model can be flexibly combined with different auxiliary information and we show the utilization of style and user intent can further enhance coherence.

There still exists a gap between our model and human poets, which indicates that there are lots to do in the future. Though we experimented on Chinese corpus, the proposed model is genre-free. We also plan to extend our model to generate other types of poetry, such as Chinese Regulated Verse and English sonnet. Besides, we also want to design some explicit supervisory signals or utilize external knowledge to improve the saliency selection algorithm.

\section*{Acknowledgments}
We would like to thank Cheng Yang, Jiannan Liang, Zhipeng Guo, Huimin Chen, Wenhao Li and anonymous reviewers for their insightful comments. This research is funded by Major Project of the National Social Science Foundation of China (No. 13\&ZD190). It is also partially supported by the NExT++ project, the National Research Foundation, Prime Minister’s Office, Singapore under its IRC@Singapore Funding Initiative.

\bibliography{conll2018}

\begin{thebibliography}{27}
\expandafter\ifx\csname natexlab\endcsname\relax\def\natexlab#1{#1}\fi

\bibitem[{Bahdanau et~al.(2015)Bahdanau, Cho, and Bengio}]{Bahdanau:15}
Dzmitry Bahdanau, KyungHyun Cho, and Yoshua Bengio. 2015.
\newblock Neural machine translation by jointly learning to align and
  translate.
\newblock In \emph{Proceedings of the 2015 International Conference on Learning
  Representations}, San Diego, CA.

\bibitem[{Blei et~al.(2003)Blei, Ng, and Jordan}]{Blei:03}
David Blei, Andrew Ng, and Michael Jordan. 2003.
\newblock Latent dirichlet allocation.
\newblock \emph{machine Learning research}, (3):993--1022.

\bibitem[{Gerv{\'a}s(2001)}]{Gervas:01}
Pablo Gerv{\'a}s. 2001.
\newblock \emph{An Expert System for the Composition of Formal Spanish Poetry}.
  Springer London.

\bibitem[{Ghazvininejad et~al.(2017)Ghazvininejad, Shi, Priyadarshi, and
  Knight}]{Ghazvininejad:17}
Marjan Ghazvininejad, Xing Shi, Jay Priyadarshi, and Kevin Knight. 2017.
\newblock Hafez: an interactive poetry generation system.
\newblock In \emph{Proceedings of ACL 2017, System Demonstrations}, pages
  43--48. Association for Computational Linguistics.

\bibitem[{Goodfellow et~al.(2013)Goodfellow, Warde-Farley, Mirza, Courville,
  and Bengio}]{Goodfellow:13}
Ian~J. Goodfellow, David Warde-Farley, Mehdi Mirza, Aaron Courville, and Yoshua
  Bengio. 2013.
\newblock Maxout networks.
\newblock In \emph{Proceedings of the 30th International Conference on Machine
  Learning}, pages 1319--1327, Atlanta, USA.

\bibitem[{He et~al.(2012)He, Zhou, and Jiang}]{He:12}
Jing He, Ming Zhou, and Long Jiang. 2012.
\newblock Generating chinese classical poems with statistical machine
  translation models.
\newblock In \emph{Proceedings of the 26th AAAI Conference on Artificial
  Intelligence}, pages 1650--1656, Toronto, Canada.

\bibitem[{Hochreiter and Schmidhuber(1997)}]{Hochreiter:97}
Sepp Hochreiter and Ju{\"a}rgen Schmidhuber. 1997.
\newblock Long short-term memory.
\newblock \emph{Neural computation}, 9(8):1735--1780.

\bibitem[{Hopkins and Kiela(2017)}]{Hopkins:17}
Jack Hopkins and Douwe Kiela. 2017.
\newblock Automatically generating rhythmic verse with neural networks.
\newblock In \emph{Proceedings of the 55th Annual Meeting of the Association
  for Computational Linguistics (Volume 1: Long Papers)}, pages 168--178.
  Association for Computational Linguistics.

\bibitem[{Jean et~al.(2017)Jean, Lauly, Firat, and Cho}]{Jean:17}
Sebastien Jean, Stanislas Lauly, Orhan Firat, and Kyunghyun Cho. 2017.
\newblock Does neural machine translation benefit from larger context?
\newblock \emph{arXiv preprint arXiv:1704.05135.}

\bibitem[{Jiang and Zhou(2008)}]{Jiang:08}
Long Jiang and Ming Zhou. 2008.
\newblock Generating chinese couplets using a statistical mt approach.
\newblock In \emph{Proceedings of the 22nd International Conference on
  Computational Linguistics}, pages 377--384, Manchester, UK.

\bibitem[{Kingma and Ba(2015)}]{Kingma:14}
Diederik~P. Kingma and Jimmy~Lei Ba. 2015.
\newblock Adam: A method for stochastic optimization.

\bibitem[{Levy(2001)}]{Levy:01}
Robert~P. Levy. 2001.
\newblock A computational model of poetic creativity with neural network as
  measure of adaptive fitness.
\newblock In \emph{Proceedings of the ICCBR-01 Workshop on Creative Systems}.

\bibitem[{Manurung(2003)}]{Manurung:03}
Hisar~Maruli Manurung. 2003.
\newblock \emph{An evolutionary algorithm approach to poetry generation}.
\newblock Ph.D. thesis, University of Edinburgh.

\bibitem[{Maruf and Haffari(2017)}]{Maruf:17}
Sameen Maruf and Gholamreza Haffari. 2017.
\newblock Document context neural machine translation with memory networks.
\newblock \emph{arXiv preprint arXiv:1711.03688}.

\bibitem[{Papineni et~al.(2002)Papineni, Roukos, Ward, and Zhu}]{P02-1040}
Kishore Papineni, Salim Roukos, Todd Ward, and Wei-Jing Zhu. 2002.
\newblock Bleu: a method for automatic evaluation of machine translation.
\newblock In \emph{Proceedings of the 40th Annual Meeting of the Association
  for Computational Linguistics}.

\bibitem[{Schuster and Paliwal(1997)}]{Schuster:97}
Mike Schuster and Kuldip~K Paliwal. 1997.
\newblock Bidirectional recurrent neural networks.
\newblock \emph{IEEE Transactions on Signal Processing}, 45(11):2673--2681.

\bibitem[{Shen et~al.(2016)Shen, Cheng, He, He, Wu, Sun, and Liu}]{P16-1159}
Shiqi Shen, Yong Cheng, Zhongjun He, Wei He, Hua Wu, Maosong Sun, and Yang Liu.
  2016.
\newblock Minimum risk training for neural machine translation.
\newblock In \emph{Proceedings of the 54th Annual Meeting of the Association
  for Computational Linguistics (Volume 1: Long Papers)}, pages 1683--1692.
  Association for Computational Linguistics.

\bibitem[{Sutskever et~al.(2014)Sutskever, Vinyals, and Le}]{Sutskever:14}
Ilya Sutskever, Oriol Vinyals, and Quoc~V. Le. 2014.
\newblock Sequence to sequence learning with neural networks.
\newblock In \emph{In Advances in Neural Information Processing Systems 2014},
  Montreal, Canada.

\bibitem[{Tiedemann and Scherrer(2017)}]{Tiedemann:17}
J{\"o}rg Tiedemann and Yves Scherrer. 2017.
\newblock Neural machine translation with extended context.
\newblock In \emph{Proceedings of the Third Workshop on Discourse in Machine
  Translation}, pages 82--92, Copenhagen, Denmark.

\bibitem[{Wang et~al.(2016)Wang, He, nad Haiyang~Wu, Li, Wang, and
  Chen}]{Wang:16}
Zhe Wang, Wei He, Hua~Wu nad Haiyang~Wu, Wei Li, Haifeng Wang, and Enhong Chen.
  2016.
\newblock Chinese poetry generation with planning based neural network.
\newblock In \emph{Proceedings of COLING 2016, the 26th International
  Conference on Computational Linguistics:Technical Papers}, pages 1051--1060,
  Osaka, Japan.

\bibitem[{Wu et~al.(2009)Wu, Tosa, and Nakatsu}]{Wu:09}
Xiaofeng Wu, Naoko Tosa, and Ryohei Nakatsu. 2009.
\newblock New hitch haiku: An interactive renku poem composition supporting
  tool applied for sightseeing navigation system.
\newblock In \emph{Proceedings of the 8th International Conference on
  Entertainment Computing}, pages 191--196, Paris, France.

\bibitem[{Yan(2016)}]{Yan:16}
Rui Yan. 2016.
\newblock i,poet:automatic poetry composition through recurrent neural networks
  with iterative polishing schema.
\newblock In \emph{Proceedings of the Twenty-Fifth International Joint
  Conference on Artificial Intelligence}, pages 2238--2244, New York, USA.

\bibitem[{Yan et~al.(2013)Yan, Jiang, Lapata, Lin, Lv, and Li}]{Yan:13}
Rui Yan, Han Jiang, Mirella Lapata, Shou-De Lin, Xueqiang Lv, and Xiaoming Li.
  2013.
\newblock I, poet:automatic chinese poetry composition through a generative
  summarization framework under constrained optimization.
\newblock In \emph{Proceedings of the 23rd International Joint Conference on
  Artificial Intelligence}, pages 2197--2203, Beijing, China.

\bibitem[{Yi et~al.(2017)Yi, Li, and Sun}]{Yi:17}
Xiaoyuan Yi, Ruoyu Li, and Maosong Sun. 2017.
\newblock Generating chinese plassical poems with rnn encoder-decoder.
\newblock In \emph{Proceedings of the Sixteenth Chinese Computational
  Linguistics}, pages 211--223, Nanjing, China.

\bibitem[{Zhang et~al.(2017)Zhang, Feng, Wang, Wang, Abel, Zhang, and
  Zhang}]{Zhang:17}
Jiyuan Zhang, Yang Feng, Dong Wang, Yang Wang, Andrew Abel, Shiyue Zhang, and
  Andi Zhang. 2017.
\newblock Flexible and creative chinese poetry generation using neural memory.
\newblock In \emph{Proceedings of the 55th Annual Meeting of the Association
  for Computational Linguistics (Volume 1: Long Papers)}, pages 1364--1373.
  Association for Computational Linguistics.

\bibitem[{Zhang and Lapata(2014)}]{Zhang:14}
Xingxing Zhang and Mirella Lapata. 2014.
\newblock Chinese poetry generation with recurrent neural networks.
\newblock In \emph{Proceedings of the 2014 Conference on Empirical Methods in
  Natural Language Processing}, pages 670--680, Doha, Qatar.

\bibitem[{Zhang(2015)}]{Zhangy:15}
Yingzhong Zhang. 2015.
\newblock \emph{How to Create Chinese Classical Poetry}. The Commercial Press.

\end{thebibliography}
\bibliographystyle{acl_natbib_nourl}
\end{document}